\documentclass[a4paper, 10pt, conference]{ieeeconf}

\usepackage{FG2019}

\FGfinalcopy

\IEEEoverridecommandlockouts                              
\def\FGPaperID{24} 

\usepackage{amssymb}
\usepackage{graphicx}
\usepackage{amsmath} 
\usepackage{subcaption}

\usepackage{color}
\usepackage{multirow}
\usepackage{booktabs}

\title{\LARGE \bf
The Deeper, the Better: Analysis of Person Attributes Recognition
}

\author{\parbox{16cm}{\centering
    {\large Esube Bekele$^1$ and Wallace Lawson$^2$}\\
    {\normalsize
    $^1$NRC Postdoctoral Fellow, Washington, DC\\
    $^2$ Naval Center for Applied Research in Artificial Intelligence, Naval Research Laboratory, Washington, DC}}
}

\begin{document}

\ifFGfinal
\thispagestyle{empty}
\pagestyle{empty}
\else
\author{Anonymous FG 2019 submission\\ Paper ID \FGPaperID \\}
\pagestyle{plain}
\fi
\maketitle

\begin{abstract}
In person attributes recognition, we describe a person in terms of their appearance.  Typically, this includes a wide range of traits including age, gender, clothing, and footwear.  Although this could be used in a wide variety of scenarios, it generally is applied to video surveillance, where attribute recognition is impacted by low resolution, and other issues such as variable pose, occlusion and shadow.  Recent approaches have used deep convolutional neural networks (CNNs) to improve the accuracy in person attribute recognition.  However, many of these networks are relatively shallow and it is unclear to what extent they use contextual cues to improve classification accuracy.  In this paper, we propose deeper methods for person attribute recognition.  Interpreting the reasons behind the classification is highly important, as it can provide insight into how the classifier is making decisions.  Interpretation suggests that deeper networks generally take more contextual information into consideration, which helps improve classification accuracy and generalizability.  We present experimental analysis and results for whole body attributes using the PA-100K and PETA datasets and facial attributes using the CelebA dataset.
\end{abstract}


\section{Introduction}
\label{intro}

Pedestrian attributes recognition involves describing the appearance of a person observed in surveillance or similar scenarios.  This description includes inherent traits such as age and gender as well as less permanent traits such as clothing, accessories and footwear.  Although it could be used in a wide range of scenarios, typically person attributes are used in person re-identification.  Re-identification can proceed in several directions.  One way is to combine many attributes together to re-identify a person \cite{martinho2018super} or by combining attributes with an another classifier \cite{schumann2017person}.

Early work on pedestrian attribute recognition proposed recognizing a small number of attributes (e.g., gender) using hand crafted features.  Unfortunately, it is difficult to hand-craft a feature that is effective for a diverse set of attributes.  For example, features that are effective for recognizing gender may not work as well for recognizing footwear.  This is the primary reason why end-to-end deep learning has become increasingly popular for attribute recognition and person re-identification.  However, most deep learning approaches for person attribute recognition use shallow CNNs (i.e., 8-layer AlexNet architecture pre-trained with ImageNet) \cite{krizhevsky2012imagenet,sudowe2015person,li2015multi,zhu2015multi}).  Later approaches moved towards deeper ResNet architectures both with\cite{yu2016weakly,sarfraz2017deep} and without pre-training \cite{bekele2017multi}, but even these are comparably shallow.  

Shallow networks are not ideal for this problem because they have a limited ability to utilize the rich amount of information present in these images.  In this paper, we demonstrate this through experimental validation on popular attributes datasets including Celeb-A, PETA, and PA-100K.  We discuss the challenges of training such deeper networks on a limited amount of training data.  Using the proposed techniques, we achieve state-of-the-art results on these datasets. We also present an interpretation with GradCAM that suggests that shallower networks typically consider less contextual information when making a classification decision.

\section{Related Work}

Early approaches to person attribute recognition involved heavy use of hand-crafted features.  Features included color histograms \cite{layne2012person, martinson2013identifying}, HOG, textures and ensemble of localized features (ELF) \cite{layne2014attributes}.  A common formula was to use a linear support vector machine (SVM) combined with hand-crafted features to recognize attributes \cite{layne2014attributes}.  

This suffers from several problems.  First, the SVM optimizes each attribute independently and lacks a way to learn relationship among attributes.  For example, attributes such as 'casual lower body' are closely related to other attributes such as 'wearing jeans', yet the SVM has no method of learning this relationship.  We can improve the performance of both by leveraging a classifier that can learn both of these attributes simultaneously.  Second, a large class imbalance can make this an extremely difficult problem as the hyperplanes are overwhelmed by a large number of negative examples compared to the rare positive examples.  Finally, it it is difficult to craft a set of features that perform well on such a wide variety of attributes in a wide variety of situations. 

Later work \cite{zhang2014panda,zhu2015multi} proposed using CNNs which showed that end-to-end learning (i.e., learning both feature and classification using stochastic gradient descent) could mitigate some of the limitations associated with support vector machines and hand-crafted features.  This has three primary benefits.  First, features are extracted using convolutional filters learned directly from the training data.  This eliminates the need to hand-craft features for each dataset and attribute. Second, the feature extractors and the classifier parameters are optimized together in an end-to-end fashion.  The extracted features are optimized for the particular attribute automatically. Finally, multi-label CNNs \cite{zhu2015multi} can significantly outperform SVMs because of their capacity to learn relationships among attributes. 

Stacked conventional CNNs have limitations in their generalization ability when trained with smaller datasets.  Because of this, pedestrian attribute recognition datasets are harder for CNNs to learn due to their limited size. The multi-label CNN (ML-CNN) in \cite{zhu2015multi}, for instance, had to be shallow with only 3 convolutional layers to cope with the smaller sample size in the datasets. Existing CNN-based approaches are either shallow networks, limiting the learning of complex features, or they are deeper networks pre-trained on larger datasets such as ImageNet \cite{li2015multi}.  In this case, representative features are not learned from training data, as is the goal of this work.  Examples of pre-training a deeper networks includes \cite{antipov2015learned,sudowe2015person,deng2015learning,zhang2014panda}. 

CNNs with branched connections in GoogLeNet \cite{szegedy2015going} and residual mapping using "shortcut" connections were proposed to tackle the accuracy degradation problem with deep CNNs\cite{he2015deep,he2016identity}. Residual networks are composed of residual units (or blocks) that have a double convolution residual leg and a direct input-to-output identity mapping or "shortcut" connection \cite{bekele2017multi}. Stacking residual blocks of up to 1000 layers, residual networks were shown to achieve state-of-the-art results on ImageNet large scale visual recognition challenge in 2015 \cite{he2015deep}. They are also shown to accelerate the learning process and converge quicker than their ``plain'' stacked CNN counterparts \cite{szegedy2016inception}. Recent work exploited joint prediction of weakly-supervised attribute locations \cite{yu2016weakly} and view \cite{sarfraz2017deep} together with attribute recognition using a pre-trained GoogLeNet and showed state-of-the-art performance on PETA. The increased performances in these methods were attributed to the joint attribute location and view prediction as well as a complicated branching schemes. Contrast that approach with this approach, where we outperform these approaches using only deeper residual networks trained with attribute recognition. 


\section{Methodology}
\label{sec:methods}

In this section, we describe our network architecture and training 
methodology.  We go further than previous work \cite{yu2016weakly,sarfraz2017deep} by using deeper residual networks (ResNet) \cite{he2016identity} for creating a richer feature sets that could achieve the state-of-the-art performance on 
attribute recognition datasets. 

\subsection{Residual Networks (ResNet)}

\begin{figure*}
	\centering
	\includegraphics[width=90mm]{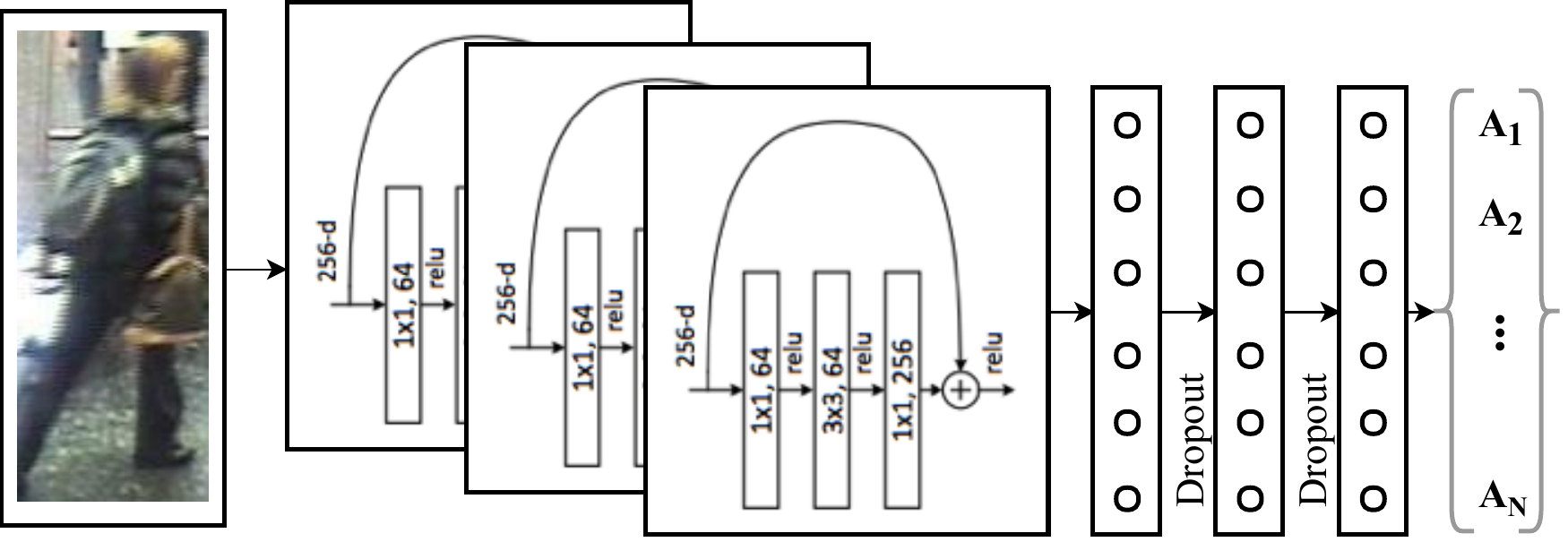}
	\includegraphics[width=60mm]{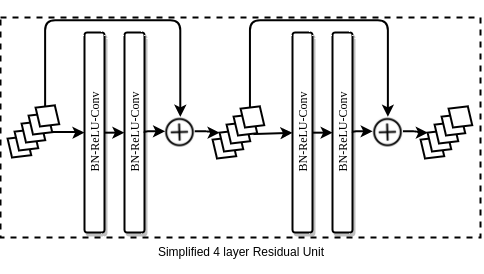}
	\caption{ResNet-based joint multi-attribute recognition model (left) and a 
	simplified residual unit (right).}
	\label{fig:arch}
\end{figure*}

The basic unit of a residual network is the residual block.  This is composed of a regular feed-forward path of stacked convolutional layers combined with a side ``short-cut'' connection. The features from both branches are merged at the end of the block.  Joint attribute learning uses stacked residual blocks as to learn relevant features.  Once an image has been processed by a number of residual blocks, the features are pooled and passed through a joint multi-label classifier (see Fig. \ref{fig:arch} right). Hence, the learned embedding at the last layer of the the feature extractor ResNet is a joint representation for all the attributes. We apply a sigmoid for the final classifier layer for jointly learning the attributes.

Previous work on person attributes recognition with relatively deeper architectures include \cite{yu2016weakly} and \cite{sarfraz2017deep} for joint person attribute learning. Both implemented variants of the GoogLeNet architecture \cite{szegedy2015going}. We explored variations of GoogLeNet, DenseNet and ResNet and found variants of ResNet performed better on joint person attributes learning (see ablation studies in the Experimental Results). We have selected ResNet as the best architecture for this problem, and use this for the remainder of the analysis.

%
%

\subsection{Feature Pooling}
After all of the residual blocks, the features are pooled to reduce dimensionality before classification.  Several methods of pooling at the feature stage have been proposed for joint multi-attribute recognition. They range from the simple global pooling at the last stage of the feature extraction pipeline to more complex pyramidal pooling schemes \cite{yu2016weakly}.

Deeper networks such as ResNet \cite{he2016identity}, GoogLeNet \cite{szegedy2015going} and DenseNet \cite{huang2016densely} pool features globally.  Global pooling reduces the dimensionality of an $h \times w \times d$ network to $1 \times 1 \times d$.  The depth of these networks gives rich features that are sufficient to be representative.  Further, they avoid the computational burden required to recognize attributes, and also avoids avoids dense fully connected classifier layer.  Unless otherwise specified, our network structure employs a global average pooling (GAP), which takes the average of each of the $d$ feature maps before performing attribute classification.

\subsection{Classifier Types}

The most common classifier (i.e., output) layers used in attribute recognition are the dense and simple logistic single output layers. Their application also depends on the depth of the network and the feature pooling method used in conjunction.  In our approach, we use a simple logistic single layer classifier.  This is preferred for deeper networks in conjunction with the global feature pooling methods.  The dense classifier is typically used for shallower networks such as DeepMAR and ACN together with local max-pooling. It is usually two fully connected layers with dropouts in between them. Because fully connected layers have exponential computational requirements with the number of neurons, they are not feasible for the rich set of features produced by deeper networks.



\subsection{Training and Optimization Objective}
We train our residual network using stochastic gradient descent with a learning rate of $0.1$.  For our loss function we use binary cross-entropy (BCE) loss\cite{zhu2015multi}.  This loss is widely used and is generally appropriate due the inherent presence of multiple attributes.  By using a multi-label BCE loss, we can not only learn about individual attributes, we can also learn about their joint interrelationship.  Equation \ref{eq:total_loss} shows BCE loss.  

\begin{equation}
\label{eq:total_loss}
Total Loss = \sum_{m=1}^{M}{\gamma_{m} loss_{m}}
\end{equation}

where $loss_{m}$ is the BCE contribution of $m^{th}$ attribute to the total loss. The $\lambda$  parameter could be used to control the learning to be focused towards a particular attribute.  This could be useful if a particular attribute is especially important for a particular recognition scenario. For instance, intrinsic attributes such as age and gender may be more important than clothing attributes for security applications. For our experiment we treated the contribution of each attribute equally by setting $\lambda$ to $\dfrac{1}{M}$.

The individual attribute losses, $loss_{m}$, are computed using the sigmoid binary cross-entropy loss for binary classification of each attribute as shown in 
Equation \ref{eqn:loss_m}. 

\begin{equation}
\begin{split}
\label{eqn:loss_m}
loss = -\dfrac{1}{N}\sum_{i=1}^{N} {\omega_{m}^{i}(y_{m}^{i}log(p(x_{m}^{i}))} \\ + {(1-y_{m}^{i})log(1-p(x_{m}^{i})))}
\end{split}
\end{equation}

where $y^{i}$ is the true attribute label for $m^{th}$ attribute and $i^{th}$ example. The predicted probabilities are computed via the sigmoid function and is given by Equation \ref{eqn:p_x}.

\begin{equation}
\label{eqn:p_x}
p(x_{m}^{i}) =  \dfrac{1}{1 + exp(-x_{m}^{i})}
\end{equation}
and $w_{m}$ is a sample weighting factor introduced to account for the bias that could occur due to inherent class imbalance within an attribute. The form of the sample weighting schemes is discussed in the next section.

This weighted multi-label optimization objective was minimized using stochastic gradient  decent (SGD) with Nestrov momentum.  

\subsection{Sample Weighting}

Sample weighting controls the bias given to each attribute.  In attribute recognition, there is typically a limited amount of highly skewed intra-class distribution.  For this reason, most attribute recognition methods employ sample-based weighting \cite{li2015multi,zhu2015multi,yu2016weakly,sarfraz2017deep}.  Unless otherwise specified, we use a the sample weighting scheme initially proposed by Li et al.\cite{li2015multi}, see Equation \ref{eqn:DeepMAR}. 

\begin{equation}
\label{eqn:DeepMAR}
\omega _{m}^{i} = \left\{\begin{matrix}
exp((1-p_{m})/\sigma^{2})\ \ if\ y_{m}^{i}=1 \\ exp(p_{m}/\sigma^{2})\ \ else
\end{matrix}\right.
\end{equation}

where $p_{m}$ is the number of positive examples in the $m^{th}$ attribute. $\sigma$ is a control parameter. This parameter can be used to control the number of true positives and hence control the recall in effect. Our goal is to maximize recognition among all attributes, and for datasets such as PETA the 35 commonly used attributes all represent a certain threshold of positive examples\cite{deng2014pedestrian}.  For that reason, unless otherwise noted, we set $\sigma=1$ for all our experiments.  A lower value would improve recall at the expense of mean accuracy.

%
%

 %

\subsection{Probability Calibration}
The output of our residual network architecture is one output neuron per attribute.  This output represents a probability that the attribute is present.  The primary challenge is developing ways to interpret this probability.  In person attribute recognition literature \cite{yu2016weakly}\cite{sarfraz2017deep}, a naive probability thresholding is commonly employed. That is, thresholding the output of the network at 0.5. This naive approach does not work as well as it does in large scale image recognition such as ImageNet.  We have found that a biased dataset skews the probabilities of each attribute.  An attribute that appears rarely will be skewed towards lower values than an attribute that appears more commonly.  This is an issue that is not typically found in popular datasets such as ImageNet, as these are all well balanced.

In this work, we propose a novel way of determining the threshold for each attribute neuron (i.e., probability calibration).  For this, we compute the ROC curve for the training dataset over each attribute and computing backwards the optimal probability threshold at the curve where precision and recall are equivalent (i.e., equal error rate). Note that these calibration parameters are computed only using the training set during training time during validation and testing stages, we used the same set of probability thresholds from the training set.  While we provide a full ablation study in the experimental results, we have found this strategy to strike a good balance between precision and recall, as it favors a strong example-based accuracy and F1 score. For this reason, we called it {\em F1 calibration}.

Probability calibration is in a sense similar in goals to sample weighting.  Sample weighting attempts to balance the weighting of the samples that are being learned by the network.  With sample weighting, naive thresholding is sensible, since the weights were balanced during training.  We can accomplish a similar result without sample weight.  In this case, we use the standard weighting of $\sigma=1$, then using probability calibration to modify the thresholds.  We further discuss this in the experimental results section.  Experimentally, we have found that probability calibration is a more flexible approach and that it typically outperforms sample weighting.


\subsection{Data Augmentation}
In this paper we apply the most commonly used data augmentation techniques in the person attribute recognition literature. These include random flip and crop, random image resizing with and without keeping the aspect ratio, mean subtraction, random RGB color jitter and random rotation. All the data augmentation in this paper was performed on-line, i.e. each image was randomly augmented by the one or combination of the listed methods in each iteration separately.

\subsection{Interpretability}

Interpretability provides important insights into the different ways that deep convolutional networks classify attributes.  For this, we use the recently proposed GradCAM approach \cite{selvaraju2017grad}.  GradCAM creates visual interpretations using the error gradients resulting from predicting each attribute.  This provides stronger responses in regions that had a greater impact in predicting a particular attribute.  This evaluation of GradCAM to the joint attribute recognition residual network is quite different from other networks that use this to evaluate softmax classifiers.  Each attribute is strongly related to other attributes, i.e. the learned embedding space is a joint space for the attributes and hence is not easier to disentangle for a single attribute.  For this reason, we expect to see this interrelationship or correlation between some attributes to manifest in the attribute predictions.  For example, a casual lower body likely will commonly appear with a casual upper body.  Therefore, the interpretation of 'casual lower body' should also highlight the relevant regions of the upper body.  On the other hand, an attribute like `wearing jeans' only impacts the lower body, so the interpretation should respond highly only in this region.

\section{Experimental Setup}
\label{sec:experiments}

\subsection{Datasets}
Attribute recognition is greatly impacted by the distance from the camera, variable lighting (e.g., soft shadow, hard shadows), pose, and severe occlusions due to other objects such as shrubs or benches\cite{shi2015transferring}.  It also must classify attributes using a very small number of training examples.  Unlike ImageNet (common dataset with typical size to train deeper models), which has 1.2 million training images, the training set for PETA has only 9,500 images.  Moreover, again unlike ImageNet, which has more or less 
equally sized images with aspect ratios close to 1, images in PETA came from a wide variety of sources with wildly different image sizes and their aspect ratio is far from 1 as the person length is mostly far larger than the width. 

Furthermore, attributes datasets such as PETA \cite{deng2014pedestrian} are challenging because of their intra-class variations and their skewed class distributions. In some attributes, the number of positive examples is smaller than 1\% of the training set. Moreover, some attributes occupy a very small area compared to the whole image. For instance, logos and special marks on t-shirts and v-neck shirts could occupy only a few pixels in area. Finally, attribute recognition is inherently a multi-label learning problem. Multiple person attributes may be present in an example image and there is inherent relationships among attributes \cite{zhu2015multi,bekele2017multi}.

Three datasets were employed to demonstrate the effectiveness of deeper feature extraction pipeline compared to complicated branching and computationally expensive series of dense layers. We demonstrate state-of-the-art performance on two person datasets:  the PEdesTrian Attribute (PETA) recognition dataset  \cite{deng2014pedestrian} and PA-100K dataset \cite{liu2017hydraplus}.  We also show recognition of attributes from the face using the celebrity faces attributes dataset (CelebA) \cite{liu2015deep}. PETA contains 19,000 images captured by surveillance cameras in indoor and outdoor scenarios. Originally the images were labeled with 61 binary attributes, which are related to age, clothing, carrying items, wearing accessories etc. There are also 4 multi-class attributes related to color. Deng et al. \cite{deng2014pedestrian} suggested to use 35 attributes due to severe class imbalance issues in the remaining. Therefore, we adopted these 35 attributes PETA images exhibit wide range of variations in illumination, pose, blurriness, resolution, background and occlusion. We evaluate with the suggested \cite{deng2014pedestrian} random train/validation/test split of 9500/1900/7600 images for equivalent comparison. The PA-100K contains 100,000 images captured from 598 scenes. This is by far the largest full body person attributes recognition dataset. Each image is labeled using 26 binary full body attributes. It also features a wide range of image resolutions than other datasets. CelebA contains 202,600 images labeled with 40 face-based attributes. We used the provided splits for both PA-100K and CelebA datasets. For the ablation analysis part, we consistently used PETA for uniform comparisons across all techniques.

\subsection{Evaluation Metrics}
Following the evaluation metrics of \cite{bekele2017multi}, the performance metrics we employed are: mean average accuracy (mA), area under the curve (computed as averaged over all samples and attributes) (AUC), example based metrics of accuracy (Acc) (i.e., percentage where all attributes were correctly classified), precision (Prec), recall (Rec), and F1 score. These metrics display a balanced performance across the attributes.

\subsection{Implementation Details}
The networks were trained with similar parameters for fair comparison. The input batch size were 16 for all the experiments. The images were mean subtracted with PETA mean pixel values. The cost function defined in Equation ~\ref{eq:total_loss} was optimized jointly for all the 35 attributes using SGD optimization with learning rate schedule that starts out at 0.1 and dropped by a factor of 10 at epochs 180 and 300. The layers were highly regularized with an $l_{2}$ regularizers of value 0.0001 to prevent over-fitting. 

\section{Results and Discussion}
\label{sec:results}

In this section, we, first, present state-of-the-art results on the three person attributes datasets. We then go deeper into analysis of the several factors that made it possible to train such deeper models with very limited and large multi-label space with a series of ablation experiments using the PETA dataset.
 

\subsection{Attribute Recognition Performance on PETA}

\begin{table}
	\caption{Comparison of multi-attribute recognition performance of proposed 
		model and the state-of-the-art on PETA. The data augmentation employed 
		are 
		mean subtraction, random color jitter, and random flip.}
	\label{tbl:sota}
	\centering
	\resizebox{\columnwidth}{!}{
		\begin{tabular}{@{}lcccccc@{}}
			\toprule
			\multicolumn{1}{c}{Networks/Method} & mA & Acc & Prec & Rec & F1 & 
			\multicolumn{1}{l}{\begin{tabular}[c]{@{}l@{}}AUC \\ (micro)\end{tabular}} 
			\\ \midrule
			MAResNet \cite{bekele2017multi} & 75.43 & - & - & 70.83 & - & - 
			\\ 
			ResNet-152 \cite{martinho2018super} & 81.65 & - & - & - & - & - \\ 
			
			ACN\cite{sudowe2015person} & 81.15 & 73.66 & 84.06 & 81.26 & 82.64 & - \\
			DeepMAR \cite{li2015multi} & 82.89 & 75.07 & 83.68 & 83.14 & 83.41 & - \\
			WPAL-GMP  \cite{yu2016weakly} & \textbf{85.50} & 76.98 & 84.07 & 85.78 & 84.90 & - \\
			VeSPA \cite{sarfraz2017deep} & 83.45 & 77.73 & 86.18 & 84.81 & 85.49 & - \\ 
			\midrule
			
			\begin{tabular}[c]{@{}l@{}}Proposed ResNet18\\ F1 calib + data 
				aug\end{tabular} 
			& 74.09 & 67.45 & 75.40 & 81.01 & 78.10 & 84.92 \\
			
			\begin{tabular}[c]{@{}l@{}}Proposed ResNet34\\ F1 calib + data 
				aug\end{tabular} 
			& 75.75 & 69.40 & 77.03 & 82.24 & 79.55 & 85.90 \\
			
			\begin{tabular}[c]{@{}l@{}}Proposed ResNet50\\ F1 calib/no data 
				aug\end{tabular} 
			& 84.18 & 78.03 & 85.78 & 85.69 & 85.73 & 90.55 \\
			\begin{tabular}[c]{@{}l@{}}Proposed ResNet50\\ F1 calib + data 
				aug\end{tabular} & 84.68 & \textbf{78.89} & \textbf{86.38} & \textbf{86.41} 
			& \textbf{86.39} & \textbf{90.96} \\ \bottomrule
		\end{tabular}
	}
\end{table}

Table \ref{tbl:sota} compares the performance of our proposed approach against 
other state-of-the-art attribute recognition approaches on the PETA dataset.  
ResNet50 with data augmentation techniques such as random flip and RGB color 
jitter outperforms the state-of-the-art in all the metrics presented with the 
exception of mean accuracy.  However, it is clear from other metrics such as 
precision, recall and F1 score that ResNet50 is a well balanced recognition 
approach performing well across a larger number of attributes.  Without any 
data augmentation, ResNet50 still has a performance that is comparable to or better 
than other approaches.

\subsection{Attribute Recognition Performance on PA-100K}
\begin{table}[b]
	\caption{Comparison of multi-attribute recognition performance of the 
	proposed model 
		and the state-of-the-art on PA-100K. The data augmentation employed are 
		mean subtraction, random color jitter, and random flip.}
	\label{tbl:sota_pa100}
	\centering
	\resizebox{\columnwidth}{!}{
		\begin{tabular}{@{}lllllll@{}}
			\toprule
			Networks/Method & mA & Acc & Prec & Rec & F1 & AUC \\ \midrule
			HydraPlusNet \cite{liu2017hydraplus} & 72.70 & 70.39 & 82.24 & 80.42 & 
			81.32 & \multicolumn{1}{c}{-} \\
			DeepMAR \cite{liu2017hydraplus} & 74.21 & 72.19 & 82.97 & 82.09 & 82.53 & 
			\multicolumn{1}{c}{-} \\ \midrule
			
			\begin{tabular}[c]{@{}l@{}}Proposed ResNet18\\ F1 calibration\end{tabular} 
			& 74.45 & 72.42 & 83.90 & 82.03 & 82.95 & 88.97 \\
			
			\begin{tabular}[c]{@{}l@{}}Proposed ResNet34\\ F1 calibration\end{tabular} 
			& \textbf{78.77} & \textbf{75.05} & \textbf{85.01} & \textbf{84.56} & 
			\textbf{84.78} & \textbf{89.95} \\
			\begin{tabular}[c]{@{}l@{}}Proposed ResNet50\\ deepmar\end{tabular} & 78.07 
			& 73.51 & 83.74 & 83.45 & 83.59 & 89.19 \\
			\begin{tabular}[c]{@{}l@{}}Proposed ResNet50\\ F1 calibration\end{tabular} & 78.12 & 74.11 & 84.42 & 84.09 & 84.25 & 89.54 \\ \bottomrule
		\end{tabular}
			
	}
\end{table}

Table \ref{tbl:sota_pa100} compares the performance of our proposed approach 
against 
other state-of-the-art attribute recognition approaches on the PA-100K 
dataset.  
Our proposed model outperforms the state-of-the-art in this dataset by about 
3\% on each metric.  From the metrics such as precision, recall and F1 score we 
can see that our model is a well balanced recognition approach performing well 
across a larger number of attributes.  In this case, we can also see that F1 calibration and
sample weighting both outperform the state-of-the-art results on this dataset.  It is also 
noteworthy that on this dataset probability calibration using F1 calibration outperforms naive 
thresholding with sample weighting.

\subsection{Attribute Recognition Performance on CelebA}
\begin{table}
	\caption{Comparison of multi-attribute recognition performance of the 
		proposed model and the state-of-the-art on CelebA. The data 
		augmentation 
		employed are mean subtraction, random color jitter, and random flip.}
	\label{tbl:sota_celeba}
	\centering
	\resizebox{\columnwidth}{!}{
		\begin{tabular}{@{}lllllll@{}}
			\toprule
			Networks/Method & mA & Acc & Prec & Rec & F1 & AUC \\ \midrule
			LNets+ANet \cite{liu2015deep} & \textbf{87} & \multicolumn{1}{c}{-} &
			\multicolumn{1}{c}{-} & \multicolumn{1}{c}{-} & \multicolumn{1}{c}{-} &
			\multicolumn{1}{c}{-} \\
			FaceTracker \cite{liu2015deep} & 81 & \multicolumn{1}{c}{-} &
			\multicolumn{1}{c}{-} & \multicolumn{1}{c}{-} & \multicolumn{1}{c}{-} &
			\multicolumn{1}{c}{-} \\
			PANDA-l \cite{liu2015deep} & 85 & \multicolumn{1}{c}{-} & \multicolumn{1}{c}{-}
			& \multicolumn{1}{c}{-} & \multicolumn{1}{c}{-} & \multicolumn{1}{c}{-} \\
			\midrule
			
			\begin{tabular}[c]{@{}l@{}}Proposed ResNet18\\ FPR@10\% 
				calibration\end{tabular} 
			& 85.75 & 63.47 & 86.43 & 70.32 & 77.55 & 88.15 \\
			
			\begin{tabular}[c]{@{}l@{}}Proposed ResNet34\\ FPR@10\% 
				calibration\end{tabular} & 86.55 & 65.18 & \textbf{86.55} & 72.34 & 78.81 & 
			\textbf{88.79} \\
			\begin{tabular}[c]{@{}l@{}}Proposed ResNet34\\ F1 calibration\end{tabular} &
			83.16 & 65.20 & 77.80 & \textbf{80.19} & 78.98 & 86.42 \\
			\begin{tabular}[c]{@{}l@{}}Proposed ResNet34\\ FPR@8\% calibration\end{tabular}
			& 85.72 & \textbf{66.67} & 82.25 & 77.79 & \textbf{79.96} & 87.98  \\
			\bottomrule
		\end{tabular}
		
	}
\end{table}

Table \ref{tbl:sota_celeba} compares the performance of our proposed model to 
that of the state-of-the-art attribute recognition approaches on the CelebA face attributes
dataset. Our approach performed competitively to the state-of-the-art 
approaches on the mean accuracy metric. The other approaches only reported the 
mean accuracy. Hence, it is not clear how balanced their recognition on the 
other metrics were. To illustrate this point, we run the same network with 
different probability calibration techniques discussed in the methods section 
and showed that it is possible to get a mean accuracy competitive to the 
state-of-the-art at the expense of precision (i.e. proposed system with 
FPR@10\% calibration). However, we would like to point out that the more 
balanced recognition performance is achieved with the F1 score calibration although the mean 
accuracy is relatively lower.

\subsection{Ablation on Effects of Training Strategies}

Here, we explore the benefit of the strategies outlined above.  We show comparisons of different strategies based on ResNet34 as that allows fair comparisons with other types of networks such as inception (GoogLeNet with 22 layers) and AlexNet-based DeepMAR (typically 8 - 12 layers) that are 
used in \cite{yu2016weakly} and \cite{sarfraz2017deep}.

\subsubsection{Architectural Comparisons}

\begin{table}[b]
	\caption{Architectural comparisons of multi-label attribute recognition  on 
		PETA.}
	\label{tbl:arch}
	\centering
	\resizebox{\columnwidth}{!}{
		\begin{tabular}{@{}lcccccc@{}}
			\toprule
			\multicolumn{1}{c}{Networks/Method} & mA & Acc & Rec & Prec & F1 & \multicolumn{1}{l}{\begin{tabular}[c]{@{}l@{}}AUC \\ (micro)\end{tabular}} \\ \midrule
			Inception v3 & \textbf{84.77} & 72.69 & 81.36 & 81.43 & 81.39 & 87.69 \\
			DenseNet34 & 81.91 & 67.76 & 78.17 & 78.18 & 78.17 & 85.56 \\
			DesneNet121 & 83.19 & 69.57 & 79.35 & 79.40 & 79.38 & 86.35 \\ \midrule
			ResNet34 & 84.54 & \textbf{72.94} & \textbf{81.57} & \textbf{81.73} & \textbf{81.65} & \textbf{87.81} \\ \bottomrule
		\end{tabular}	
	}
\end{table}

We compare three deeper architectures: Inception version 3 (a variant 
of the 22-layer GoogLeNet-based architecture), DenseNet34 (with 34 layers), 
DenseNet121 (121 layers), and ResNet34. Table \ref{tbl:arch} shows the benefits 
of residual blocks.  ResNet34 is comparable to, but slightly outperforms, GoogLeNet 
(Inception v3).  ResNet34 also outperforms both variants of DenseNet (34 and 121 layers). 
 It is important to note here that Inception will quickly overfit with deeper architectures, and with increased 
data augmentation it trains at least twice slower than ResNet34. As we go 
deeper in these networks without additional strategies, the networks 
overfit too quickly as shown on the performance of DenseNet34 vs. DenseNet121. The slightly higher mean average accuracy of GoogLeNet when compared to ResNet34 is due to the fact that ResNet34 is trained using F1-calibration for balanced performance on the other metrics (specifically F1 metric). 

\subsubsection{Effects of Pooling and Classifiers Types}

\begin{table}
	\caption{Effect of classifier type on ResNet34 recognition performance on 
		PETA.}
	\label{tbl:clf}
	\centering
	\resizebox{\columnwidth}{!}{
		\begin{tabular}{@{}lcccccc@{}}
			\toprule
			\multicolumn{1}{c}{Networks/Method} & mA & Acc & Rec & Prec & F1 & \multicolumn{1}{l}{\begin{tabular}[c]{@{}l@{}}AUC \\ (micro)\end{tabular}} \\ \midrule
			\begin{tabular}[c]{@{}l@{}}GAP + Logistic \\ classifier\end{tabular} & \textbf{84.59} & \textbf{72.87} & \textbf{81.39} & \textbf{81.54} & \textbf{81.46} & \textbf{87.77} \\
			\begin{tabular}[c]{@{}l@{}}GAP + Dense \\ classifier\end{tabular} & 84.35 & 71.83 & 80.71 & 80.79 & 80.78 & 87.29 \\ \bottomrule
		\end{tabular}	
	}
\end{table}

In practice, global average pooling (GAP) works as well as or better than complicated 
pooling strategies. As shown on Table \ref{tbl:clf}, the simple logistic 
classifier outperforms the dense classifier for ResNet34. Generally, for 
shallower networks (ResNet34 and 
below), logistic classifier performs better than dense classifiers as the 
number of features pooled globally is limited (512 in the case of ResNet34). 
However, for deeper networks such as ResNet50 and above, the dense classifier 
works well as the number of globally pooled features is larger (2048 in the 
case of ResNet50).

\subsubsection{Sample Weights vs. Probability Calibration}

\begin{table}[b]
	\caption{Effects of sample weighting and probability calibration strategies 
		on ResNet34 performance  on PETA}.
	\label{tbl:f1}
	\centering
	\resizebox{\columnwidth}{!}{
		\begin{tabular}{@{}lcccccl@{}}
			\toprule
			\multicolumn{1}{c}{Networks/Method} & mA & Acc & Rec & Prec & F1 & \begin{tabular}[c]{@{}l@{}}AUC \\ (micro)\end{tabular} \\ \midrule
			\begin{tabular}[c]{@{}l@{}}DeepMar weights +\\ F1 calibration\end{tabular} & 84.59 & 72.87 & 81.39 & 81.54 & 81.46 & \multicolumn{1}{c}{87.77} \\
			\begin{tabular}[c]{@{}l@{}}DeepMar weights + \\ recall @20\% FPR\end{tabular} & \multicolumn{1}{l}{\textbf{92.24}} & \multicolumn{1}{l}{65.73} & \multicolumn{1}{l}{\textbf{93.19}} & \multicolumn{1}{l}{67.59} & \multicolumn{1}{l}{78.35} & 86.60 \\
			\begin{tabular}[c]{@{}l@{}}No weighting/\\ naive proba thresh\end{tabular} & \multicolumn{1}{l}{83.07} & \multicolumn{1}{l}{72.25} & \multicolumn{1}{l}{79.47} & \multicolumn{1}{l}{82.95} & \multicolumn{1}{l}{81.17} & 87.19 \\
			\begin{tabular}[c]{@{}l@{}}No weighting/\\ F1 calibration\end{tabular} & 85.33 & \textbf{73.54} & 82.09 & \textbf{82.22} & \textbf{82.16} & \multicolumn{1}{c}{\textbf{88.23}} \\ \bottomrule
		\end{tabular}	
	}
\end{table}

From Table \ref{tbl:f1}, it is clear that the naive probability threshold (thresholding at 0.5) with 
no sample weighting performed the worst. Next, the sample weighting (DeepMAR weights)
together with F1 probability calibration performs better. The F1 probability 
calibration with no sample weighting achieved the best and well balanced performance among 
these comparisons.  DeepMAR weights with a 20\% false positive rates gives a higher mean accuracy, 
but this is at the expensive of more false positives, which is measured in the other metrics.

\subsubsection{Effects of Image Resizing and Data Augmentation}

Other factors such as the way images are resized before being fed to the network can have a big impact.  Strategies such as data augmentation  and depth of the networks could affect 
performance. It is a combination of these factors together with the strategies 
discussed above that resulted in the state-of-the-art performance on PETA as 
shown in Table \ref{tbl:da} and Table \ref{tbl:sota}.   

For augmentation, we experiment with applying a random flip and RGB color jitter while preserving the aspect ratio of the original image. 
Some network inputs are rigid and do not keep the aspect ratio of the input 
\cite{li2015multi} and \cite{sarfraz2017deep}, requiring an image with an input 
size of 256x256 with crop size of 224x224. But, a dramatic performance increase 
was observed by preserving the aspect ratios of input images with random 
sampling of the largest dimension from a set. This results in an increase in 
example-based accuracy (almost 6\%) and in F1 score (about 4\%).  Intuitively, this makes sense if we think about an attribute such as a ``wearing a v-neck shirt''.  Stretching out the image can make it difficult to see this attribute, which clearly impacts performance.  

\begin{table}
	\caption{Effects of image resizing (Size and Crop) and data augmentation 
		(DA) on performance.}
	\label{tbl:da}
	\centering
	\resizebox{\columnwidth}{!}{
		\begin{tabular}{@{}lllllll@{}}
			\toprule
			\multicolumn{1}{c}{Networks/Method} & \multicolumn{1}{c}{mA} & \multicolumn{1}{c}{Acc} & \multicolumn{1}{c}{Rec} & \multicolumn{1}{c}{Prec} & \multicolumn{1}{c}{F1} & \begin{tabular}[c]{@{}l@{}}AUC \\ (micro)\end{tabular} \\ \midrule
			\begin{tabular}[c]{@{}l@{}}Size: 256x256\\ Crop: 224x224\\ DA: flip, crop\end{tabular} & \multicolumn{1}{c}{84.59} & \multicolumn{1}{c}{72.87} & \multicolumn{1}{c}{81.39} & \multicolumn{1}{c}{81.54} & \multicolumn{1}{c}{81.46} & \multicolumn{1}{c}{87.77} \\
			\begin{tabular}[c]{@{}l@{}}Size: 300x300\\ Crop: 299x299\\ DA: flip, crop\end{tabular} & 84.54 & 72.94 & 81.57 & 81.73 & 81.65 & 87.81 \\
			\begin{tabular}[c]{@{}l@{}}Size: random\\ AR: preserved\\ DA: None\end{tabular} & 84.18 & 78.03 & 85.69 & 85.78 & 85.73 & 90.55 \\
			\begin{tabular}[c]{@{}l@{}}Size: random\\ AR: preserved\\ DA: flip, jitter, rot\end{tabular} & \multicolumn{1}{c}{83.19} & \multicolumn{1}{c}{77.16} & \multicolumn{1}{c}{84.96} & \multicolumn{1}{c}{85.12} & \multicolumn{1}{c}{85.04} & \multicolumn{1}{c}{90.03} \\ \midrule
			\begin{tabular}[c]{@{}l@{}}Size: random\\ AR: preserved\\ DA: flip, jitter\end{tabular} & \textbf{84.68} & \textbf{78.89} & \textbf{86.41} & \textbf{86.38} & \textbf{86.39} & \textbf{90.96} \\ \bottomrule
		\end{tabular}	
	}
\end{table}

\subsection{Interpretability}

Recent work by \cite{selvaraju2017grad} demonstrated a method of interpreting classification results without the need to re-train networks.  This approach, known as Grad-CAM, operates by performing a backward pass through the network and pooling in order to show the regions that are responsible for classification of a particular attribute.  We use Grad-CAM to provide an insight into deeper person attribute recognition.  

\begin{figure*}
	\centering
	\begin{tabular}{lllllcccrrrr}
	\small{Original Image} & & & & & & \small{ResNet with 18 layers} & & &  & & \small{ResNet with 34 layers} \\
	\end{tabular}
	\begin{subfigure}[b]{0.85\linewidth}
	\includegraphics[width=15mm]{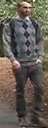}  
	\includegraphics[width=15mm]{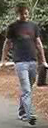}
	\includegraphics[width=15mm]{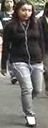}
	\hspace{1mm}
	\includegraphics[width=15mm]{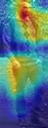}
	\includegraphics[width=15mm]{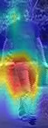}
	\includegraphics[width=15mm]{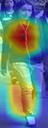}
	\hspace{1mm}
	\includegraphics[width=15mm]{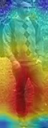}
	\includegraphics[width=15mm]{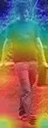}	
	\includegraphics[width=15mm]{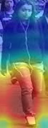}
	\caption{Interpretation of the attribute ``Wearing Jeans''}
	\end{subfigure}
	\begin{subfigure}[b]{0.85\linewidth}
	\includegraphics[width=15mm]{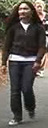} 
	\includegraphics[width=15mm]{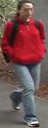}
	\includegraphics[width=15mm]{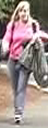}
	\hspace{1mm}
	\includegraphics[width=15mm]{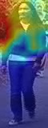}
	\includegraphics[width=15mm]{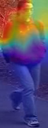}
	\includegraphics[width=15mm]{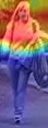}
	\hspace{1mm}
	\includegraphics[width=15mm]{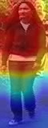}
	\includegraphics[width=15mm]{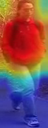}
	\includegraphics[width=15mm]{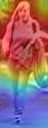}
	\caption{Interpretation of the attribute ``Long Hair''}
	\end{subfigure}
	\begin{subfigure}[b]{0.85\linewidth}
	\includegraphics[width=15mm]{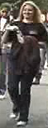} 
	\includegraphics[width=15mm]{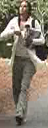}
	\includegraphics[width=15mm]{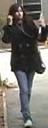}
	\hspace{1mm}
	\includegraphics[width=15mm]{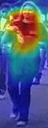}
	\includegraphics[width=15mm]{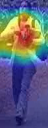}
	\includegraphics[width=15mm]{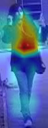}
	\hspace{1mm}
	\includegraphics[width=15mm]{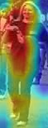}
	\includegraphics[width=15mm]{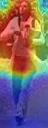}
	\includegraphics[width=15mm]{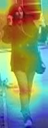}
	\caption{Interpretation of the attribute ``Carrying Messenger Bag''}
	\end{subfigure}

	\caption{An interpretation of why a ResNet18 (middle) and ResNet34 (right) classifies various attributes. }
	\label{Interpret1}
\end{figure*}

Figure \ref{Interpret1} shows the GradCAM interpretation of why attributes were correctly recognized.  The figure shows a residual network with 18 layers (ResNet18) and a residual network with 34 layers (ResNet34).  Each row shows a different attribute.  The top row shows the interpretation of recognizing  jeans (top row), long hair (middle row) and having a messenger bag (bottom row).  These are all true positives.  Experimentally, we already have shown that ResNet34 outperforms ResNet18 in terms of classification accuracy.  Here we see that ResNet34 also seems to take more of the image into consideration as well.  The top row (jeans) and bottom row (messenger bag) shows two good examples of this.  Whereas ResNet18 tries to locate small regions, ResNet34 is looking at bigger parts of the image before making a classification decision.  The middle row (long hair) shows a very hard to classify attribute.  Here, it appears that ResNet34 is looking to take contextual information (possibly gender and clothing) into consideration as well.  ResNet18 focuses only on the hair area.  
 
\begin{figure*}[t!]
	\centering

	\begin{subfigure}[b]{0.4\textwidth}
		\centering
		\includegraphics[height=30mm,width=15mm]{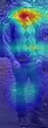}
    	\includegraphics[height=30mm,width=15mm]{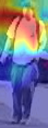}
        \includegraphics[height=30mm,width=15mm]{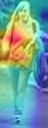}
        \caption{ResNet18 false positives.  From left to right:  \\Wearing short sleeves, Long hair, Wearing skirt}
    \end{subfigure}
    \begin{subfigure}[b]{0.4\textwidth}
        \centering
        \includegraphics[height=30mm,width=15mm]{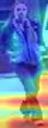}
	 	\includegraphics[height=30mm,width=15mm]{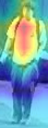}
	 	\includegraphics[height=30mm,width=15mm]{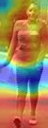}
	 	\caption{ResNet34 false positives.  From left to right: \\Male, Long Hair, and Age 31-45.}
    \end{subfigure}
	\caption{False Positives: ResNet18 (left) and ResNet34 (right) visualization using GradCAM}
	\label{Interpret2}
\end{figure*}

Figure \ref{Interpret2} shows three false positives from ResNet18 (left) and 
ResNet34 (right).  The images on the left show misclassifications of short 
sleeves, long hair, and skirt.  The images on the right show misclassifications 
of gender, long hair, and age (31-45 predicted).  We show these because in some 
cases, the interpretation is odd enough to call into question the accuracy of 
that classification.  In some cases (short sleeves, skirt) the network is not looking at 
the correct part of the image.  In other cases (gender, age) it is making predictions based on very odd patterns in the image.  

\section{Conclusion}
\label{sec:conclusion}

Recognizing attributes in the wild can be very challenging.  Environmental conditions such as weather, lighting and shadow can impact results.  Poor color calibration of the camera is also quite common.  People also can be partially occluded by other things in the environment.  Despite all of these challenges, we wish to have a network that can recognize a wide range of attributes about a person while doing so quite quickly.  

From our experiments, we draw the following conclusions.   First, residual networks are appropriate when training deeper networks with limited size datasets.  We have demonstrated that both ResNet34 and ResNet50 outperform state-of-the-art, with the later resulting in our best overall performance.  These networks perform well at the most important metric, example accuracy, showing that the network is trying to strike a balance between recognizing frequently appearing attributes such as gender, while not completely ignoring infrequently appearing attributes like v-neck shirt.  

A number of strategies helped to contribute to this balanced performance.  We found the best improvement on our accuracy to come from preserving the aspect ratio of the input image.  We hypothesize that this has a huge impact on attributes that can be distorted during the resizing process.  For example, a v-neck t-shirt becomes hard to identify if the shirt has been stretched out too far.  Data augmentation can have a slight increase in the performance of the network, but we have also found that the augmentation should well match the data collection procedure.  

Probability calibration strategies can also greatly impact the results, but care needs to be taken to ensure that the chosen strategy performs well across a wide range of attributes, and not just biasing the results towards those classes that appear most frequently.  Finally, it is important to consider the amount of time that is required to recognize the attributes of a person.  ResNet50 has some initial cost associated with loading the model and initializing weights, but once running it will process images of persons very quickly.

\section*{Acknowledgements}

Wallace Lawson was supported by the Office of Naval Research, Esube Bekele was supported by the National Research Council.

\bibliographystyle{ieee}
\bibliography{egbib}

\end{document}